\pdfoutput=1
\documentclass[11pt]{article}

\usepackage[preprint]{acl}

\usepackage{times}
\usepackage{latexsym}

\usepackage[T1]{fontenc}
\usepackage[utf8]{inputenc}

\usepackage{microtype}

\usepackage{inconsolata}

\usepackage{graphicx}
\usepackage{hyperref}

\usepackage{url}            % simple URL typesetting
\usepackage{booktabs}       % professional-quality tables
\usepackage{amsfonts}       % blackboard math symbols
\usepackage{nicefrac}       % compact symbols for 1/2, etc.
\usepackage{microtype}      % microtypography
\usepackage{xcolor}         % colors
\usepackage{tabularx}
\usepackage{times}

\usepackage{amsmath}
\usepackage{amssymb}
\usepackage{mathtools}
\usepackage{wrapfig}
\usepackage{float}
\usepackage{multirow}
\usepackage{graphicx,subcaption}

\usepackage{rotating}
\usepackage{svg}
\usepackage{enumitem}
\usepackage{bm,xspace}

\definecolor{gpts}{rgb}{0.839, 0.153, 0.157}
\definecolor{gptm}{rgb}{0.12156862745098039, 0.4666666666666667, 0.7058823529411765}
\definecolor{gptl}{rgb}{1.0, 0.4980392156862745, 0.054901960784313725}
\definecolor{gptxl}{rgb}{0.17254901960784313, 0.6274509803921569, 0.17254901960784313}
\definecolor{olmo}{rgb}{0.5, 0.0, 0.53}

\title{Adaptation Odyssey in LLMs: Why Does Additional Pretraining Sometimes Fail to Improve?}

\author{
  \textbf{Fırat Öncel\textsuperscript{1, 2}},
  \textbf{Matthias Bethge\textsuperscript{3,4}},
  \textbf{Beyza Ermis\textsuperscript{5}},
\\
  \textbf{Mirco Ravanelli\textsuperscript{1,2}},
  \textbf{Cem Subakan\textsuperscript{1,2,6}},
  \textbf{Çağatay Yıldız\textsuperscript{3,4}}
\\
\\
  \textsuperscript{1}Concordia University,
  \textsuperscript{2}Mila-Quebec AI Institute,
  \textsuperscript{3}University of Tübingen,
  \\
  \textsuperscript{4}Tübingen AI Center,
  \textsuperscript{5}Cohere For AI,
  \textsuperscript{6}Laval University
\\
  \small{
    \textbf{Correspondence:} \href{mailto:firat.oncel@mail.concordia.ca}{firat.oncel@mail.concordia.ca}
  }
}

\newcommand{\gpt}{\text{GPT2}\xspace}
\newcommand{\gpts}{\text{GPT2-small}\xspace}

\newcommand{\gptl}{\text{GPT2-large}\xspace}
\newcommand{\gptxl}{\text{GPT2-xlarge}\xspace}
\newcommand{\olmo}{\text{OLMo-1B}\xspace}
\newcommand{\llama}{\text{LLaMA-7B}\xspace}

\begin{document}
\maketitle
\begin{abstract}
In the last decade, the generalization and adaptation abilities of deep learning models were typically evaluated on fixed training and test distributions.
Contrary to traditional deep learning, large language models (LLMs) are (i) even more overparameterized, (ii) trained on unlabeled text corpora curated from the Internet with minimal human intervention, and (iii) trained in an online fashion.
These stark contrasts prevent researchers from transferring lessons learned on model generalization and adaptation in deep learning contexts to LLMs.

To this end, our short paper introduces empirical observations that aim to shed light on further training of already pretrained language models. 
Specifically, we demonstrate that training a model on a text domain could degrade its perplexity on the test portion of the same domain. 
We observe with our subsequent analysis that the performance degradation is positively correlated with the similarity between the additional and the original pretraining dataset of the LLM.
Our further token-level perplexity observations reveals that the perplexity degradation is due to a handful of tokens that are not informative about the domain. 
We hope these findings will guide us in determining when to adapt a model vs when to rely on its foundational capabilities.
\end{abstract}

\section{Introduction}
\label{sec:intro}
Deep learning generalization theory and empirical studies have traditionally assumed a fixed data distribution from which training and test datasets are sampled \citep{neyshabur2017exploring}. This train-test paradigm was later evolved by \emph{domain adaptation} and \emph{continual learning}, where the original training distribution differs from future distributions to be fitted. The advent of foundation models has marked a significant shift, as these general-purpose models are pretrained on enormous datasets, which may not even be published \citep{kaplan2020scaling}. Furthermore, many datasets are known to have data leakage, where train and test points are duplicates \cite{dolma}. Consequently, in this modern era of machine learning, the clear train-test dichotomy does not apply for LLM training. 

This short paper stems from our curiosity about whether conventional machine learning principles remain relevant amidst the aforementioned paradigm shift. Specifically, we aim to understand to what extent the deep learning optimization and generalization practices of the last decade can be applied today. Our primary question is the following:
\textit{Is it still relevant to study additional pretraining of models that have already been trained on possibly unknown text corpora by LLM engineers?}

%We believe our initial attempts to answer this question would guide the research in domain adaptation, where the initial findings somewhat conflict \citep{gururangan2020don, cheng2023adapting}. 

% c's version
The earlier works in the literature pertinent to this question have conflicting findings \citep{gururangan2020don, cheng2023adapting}. However, we believe that the empirical findings in the paper help to improve our understanding on this subject by presenting more consistent observations.

For our investigation, we adapt LLMs of various sizes and architectures to different domains within the Massively Multi-Domain Dataset (M2D2, \citep{reid2022m2d2}), a carefully curated collection of over 200 text domains from Wikipedia (Wiki) and Semantic Scholar (S2ORC). We compare the perplexities obtained on the test set of a domain before and after training on the same domain. While it is generally expected that adaptation to a new domain would improve the within-domain test perplexity, our findings suggest this is not always the case.

Interestingly, we observe that additional pretraining on Wiki domains tends to degrade test perplexity, while pretraining on S2ORC domains always improves it. To quantify this intuitive observation, we measure the distributional similarities between additional training domains and the original pretraining corpora. Our results show that the performance degradation is positively correlated with the similarity of the training domains' embeddings to those of the original pretraining set and the additional pretraining set.
We further analyze how adaptation changes the perplexity of individual tokens, and discover that most of the degradation can be attributed to a few tokens unrelated to any domain, such as ``\texttt{\textbackslash n}'', making it difficult to rely on perplexity (averaged over a test set) as a measure of improvement.

\section{Method}
\label{sec:method}
In this section we present training details, source corpora and adaptation domains details, evaluation method and domain similarity measures.
\subsection{Models and Training}
\label{sec:training}
% to extract embeddings for the analyses
We conduct our experiments with decoder-only GPT2 model family \cite{radford2019language}, such as \gpts, \gptl and \gptxl, \olmo \cite{groeneveld2024olmo} and \llama \cite{touvron2023llama} models.
We additional pretrain models on M2D2 domains separately (see the next section for details) using the Deepspeed library \cite{rasley2020deepspeed}. We use a learning rate of $0.00005$ for the \gpt models, $0.000005$  for the \llama, and $0.0000085$ for the \olmo model. We additional pretrain each model for 1 epoch on a single GPU.

Our domain similarity analyses require access to the training corpus of the said LLMs, which is why we choose open-data models. To conduct the analyses, we sample 400k texts from GPT2's training corpus, OpenWebText \citep{Gokaslan2019OpenWeb}, 650k texts from OLMo's training corpus, Dolma \citep{dolma} and 930k text from LlaMa's training corpus, \cite{together2023redpajama}.

\subsection{Tasks}
\label{sec:task}
We conduct experiments on 20 adaptation domains from M2D2 Dataset, adaptation domains are presented in Appendix \ref{sec:adaptation-domain}. Half of the domains belong to the Wiki portion, while the other half belong to the S2ORC portion of the dataset. We choose the adaptation domains based on the similarity measures explained in section \ref{sec:similarity}.
The selected S2ORC domains include: \textit{High Energy Physics}, \textit{Nuclear Experiment}, \textit{Condensed Matter}, \textit{Mathematics}, \textit{Super Conductivity} and \textit{Astrophysics} while Wiki domains include: \textit{Society and Social Sciences}, \textit{Technology and Applied Sciences}, \textit{Human Activities}, \textit{Culture and the Arts}, \textit{History and Events}, \textit{Philosophy and Thinking}, \textit{Natural and Physical Sciences} and \textit{General Reference}.
\subsection{Evaluation}
\label{sec:evaluation}
We evaluate model's perplexities on generation task with the additional pretrained model on that specific domain for a single epoch.

\subsection{Domain Similarity Measures}
\label{sec:similarity}
We use two similarity measures to compare the similarity between original corpora and adaptation domains. For each source corpus and target domain, we randomly sample 5\% of the texts for large domains or up to 50 000 texts for small domains (when feasbible). We then extract $d$-dimensional $l_2$ normalized embeddings using Sentence Transformers (SBERT) \cite{reimers-2019-sentence-bert}. We define the corpus embeddings with $M$ samples as $ \mathcal{C} =[\theta(\mathcal{C}_{t_1}), ..., \theta(\mathcal{C}_{t_M})] $, and domain embeddings with $N$ samples as $ \mathcal{D} =[\theta(\mathcal{D}_{t_1}), ..., \theta(\mathcal{D}_{t_N})] $ where $\theta$ is the feature extractor SBERT model. We calculate the following similarity measures as follows:
\paragraph{Maximum Mean Discrepancy (MMD). }

We use the closed-form expression from \cite{gretton12mmd} to calculate MMD, with linear kernel, between the source corpus  $\mathcal{C}$ and target domain $\mathcal{D}$: $MMD(\mathcal{C}, \mathcal{D}) = \left\|\mu_C-\mu_D\right\|_2^2$, where $ \mu_C $ and $ \mu_D $ are $d$-dimensional sample means.

\paragraph{Fréchet Distance (FD). }
We use the closed-form expression from \cite{frechet1982} to calculate FD between the source corpus and target domain. FD between corpus $\mathcal{C}$ and target domain $\mathcal{D}$:
$FD(\mathcal{C}, \mathcal{D})=\left\|\mu_C-\mu_D\right\|_2^2+\operatorname{tr}\left(\Sigma_C+\Sigma_D-2 \sqrt{\Sigma_C \Sigma_D}\right)$  where $ \mu_C $ and $ \mu_D $ are $d$-dimensional sample means, $ \Sigma_C $ and $ \Sigma_D $ are $(d, d)$-dimensional sample covariances.

\begin{figure*}[ht!]
	\centering
	\includegraphics[width=\linewidth]{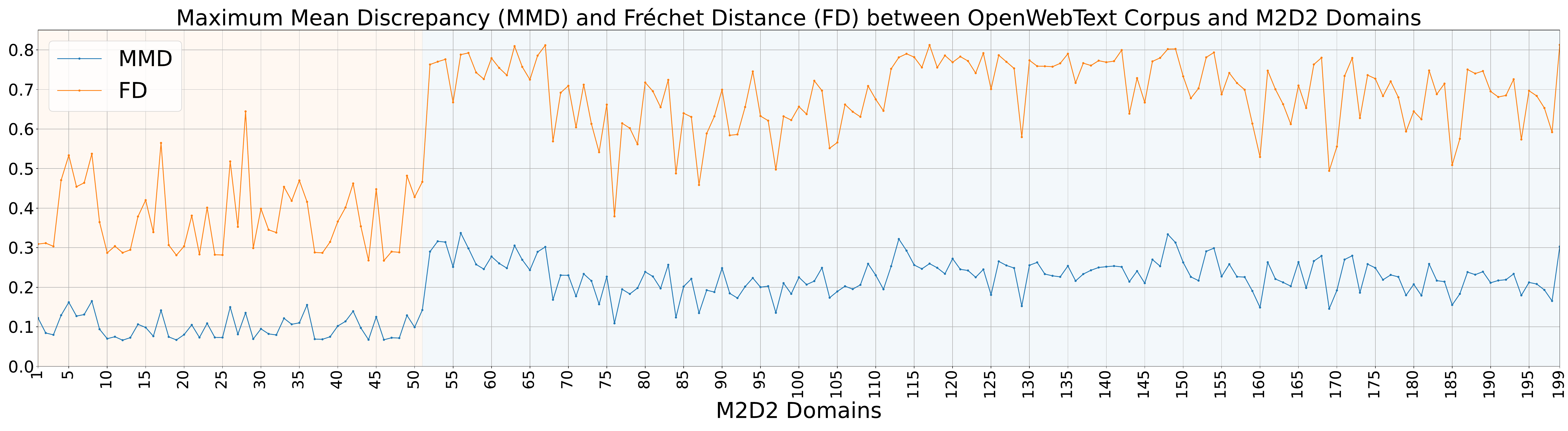}
	\caption{Domain IDs ($x$ axis). MMD and FD scores between OpenWebText and M2D2  Domains ($y$ axis). Wiki (blue shaded area) portion is closer to source corpora compared to the S2ORC (orange shaded area)  portion. All Domain names corresponding to the IDs in $x$ axies are presented in Appendix~\ref{sec:all-domain}. Same plot for Dolma is presented in Appendix, Figure~\ref{fig:mmd-fid-dolma}.}
	\label{fig:mmd-fid}
 
\end{figure*}

MMD and FD scores between original corpora and adaptation domains are presented in Figure~\ref{fig:mmd-fid}.

\begin{figure*}[ht!]
	\centering
	\includegraphics[width=.48\linewidth]{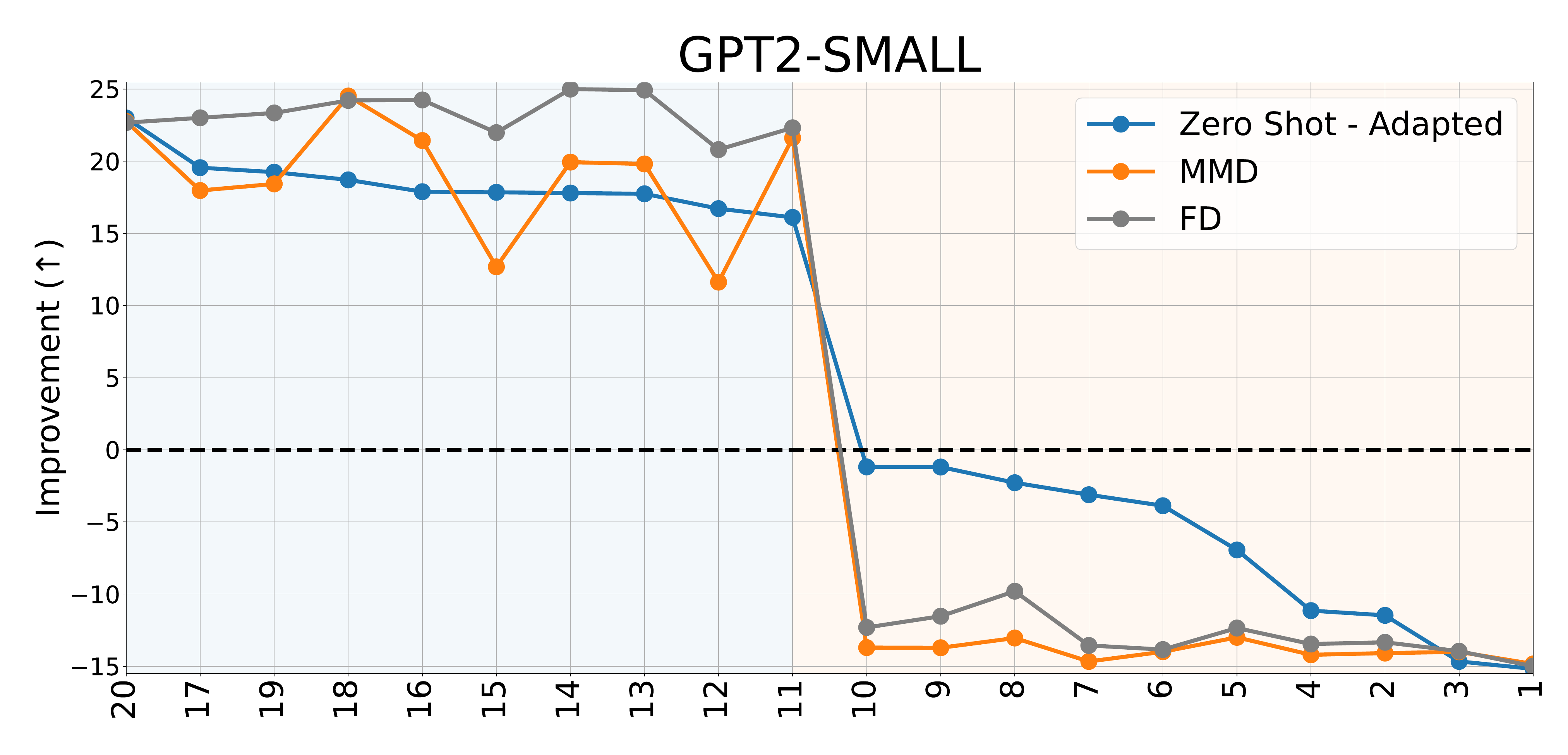}
	\includegraphics[width=.48\linewidth]{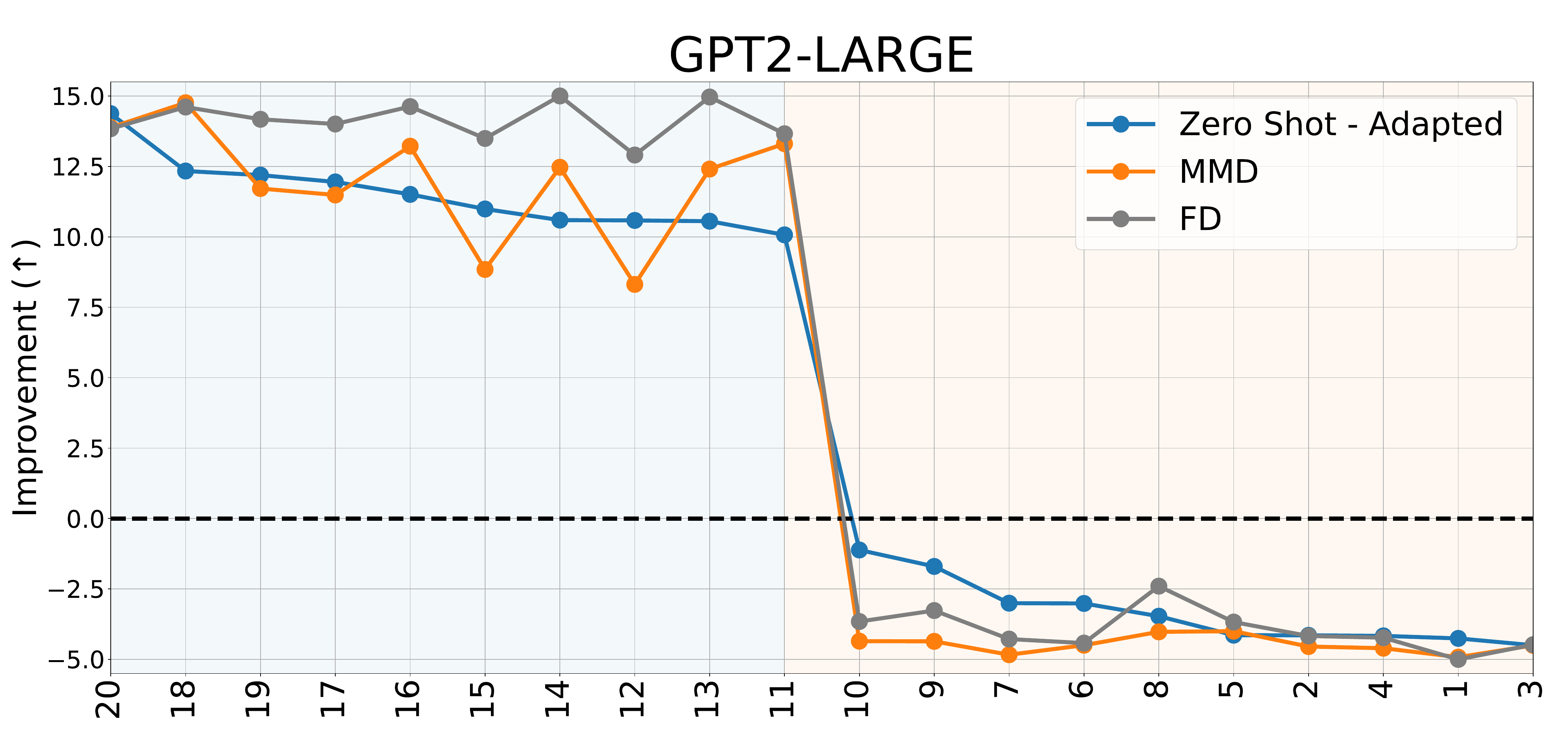}
	\includegraphics[width=.48\linewidth]{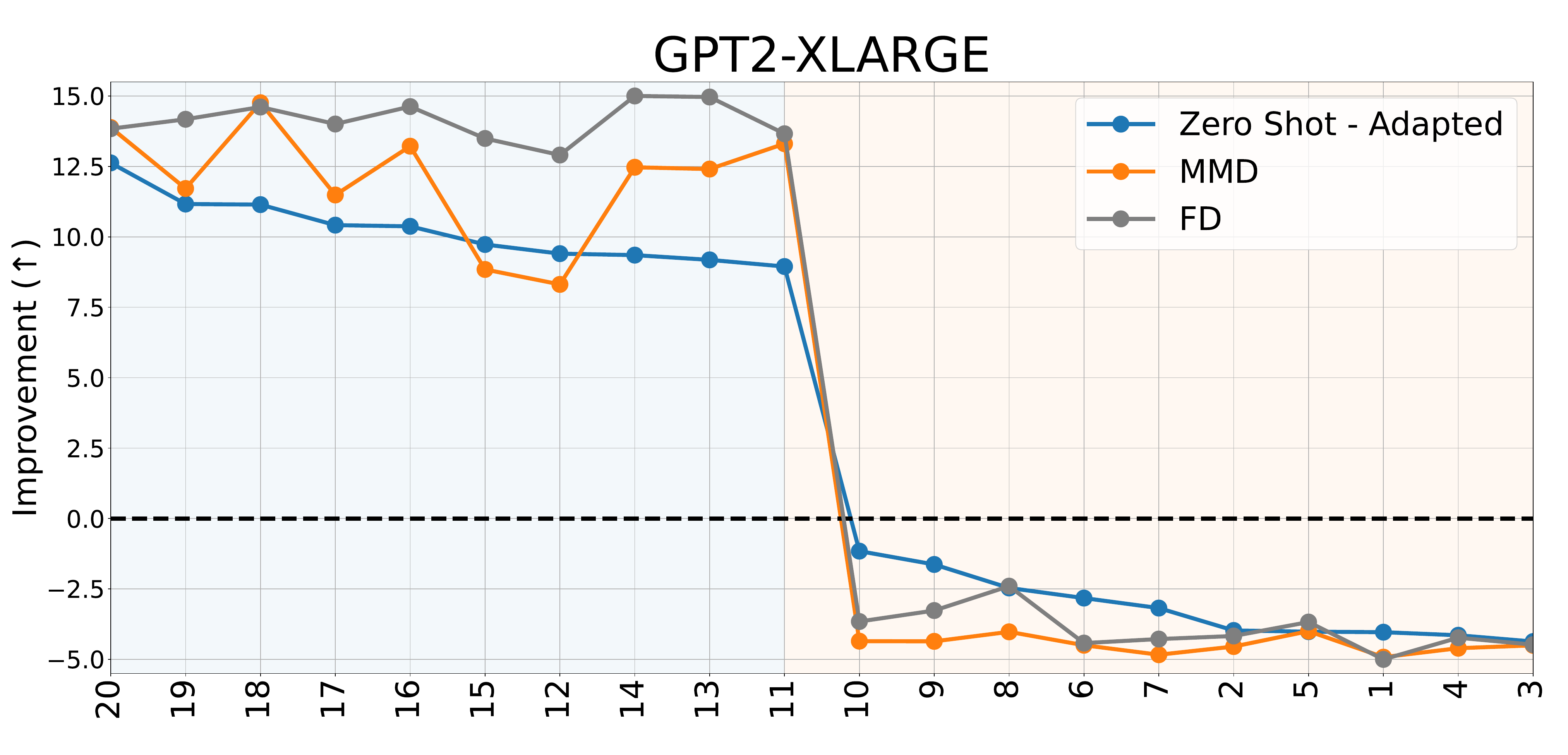}
	\includegraphics[width=.48\linewidth]{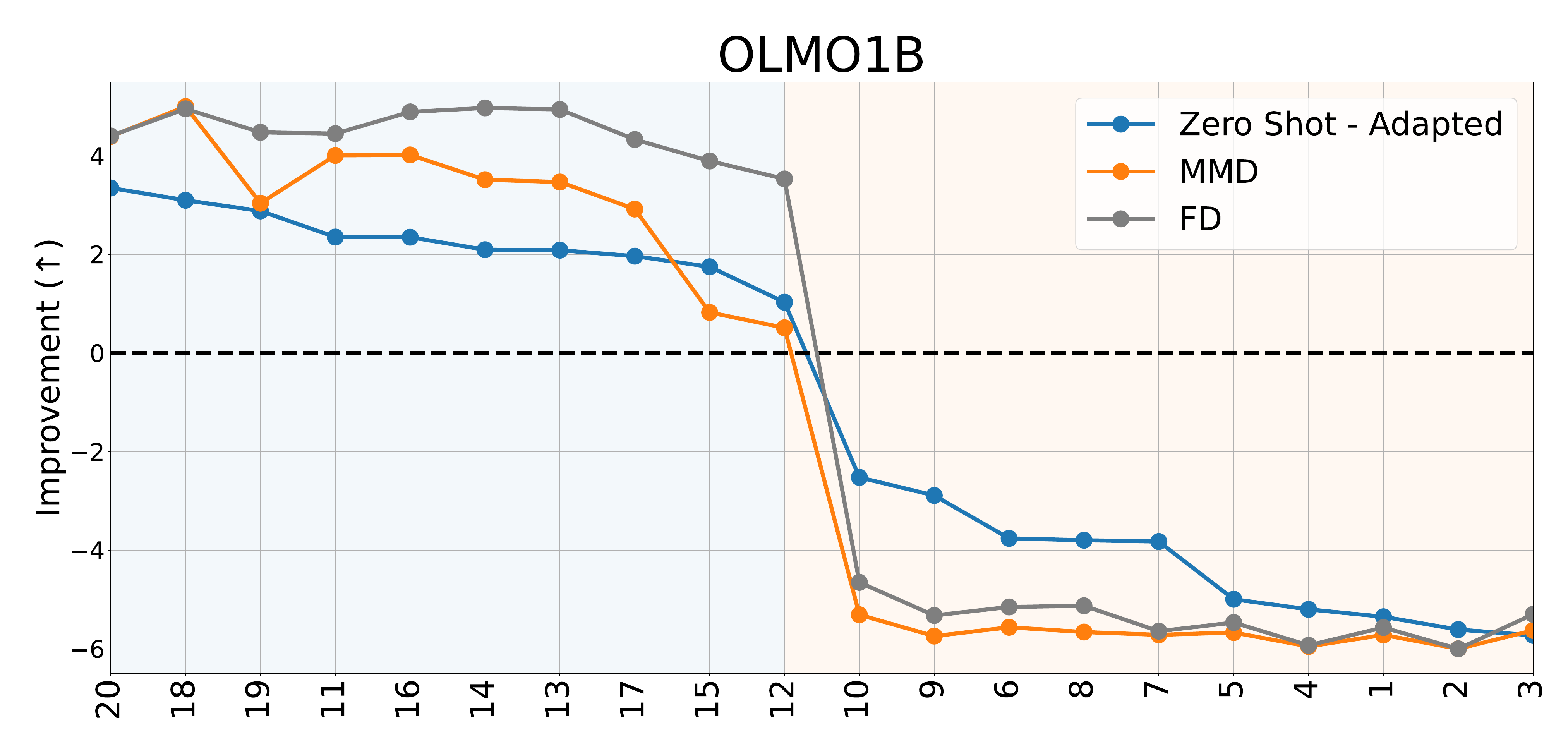}
	\includegraphics[width=0.48\linewidth]{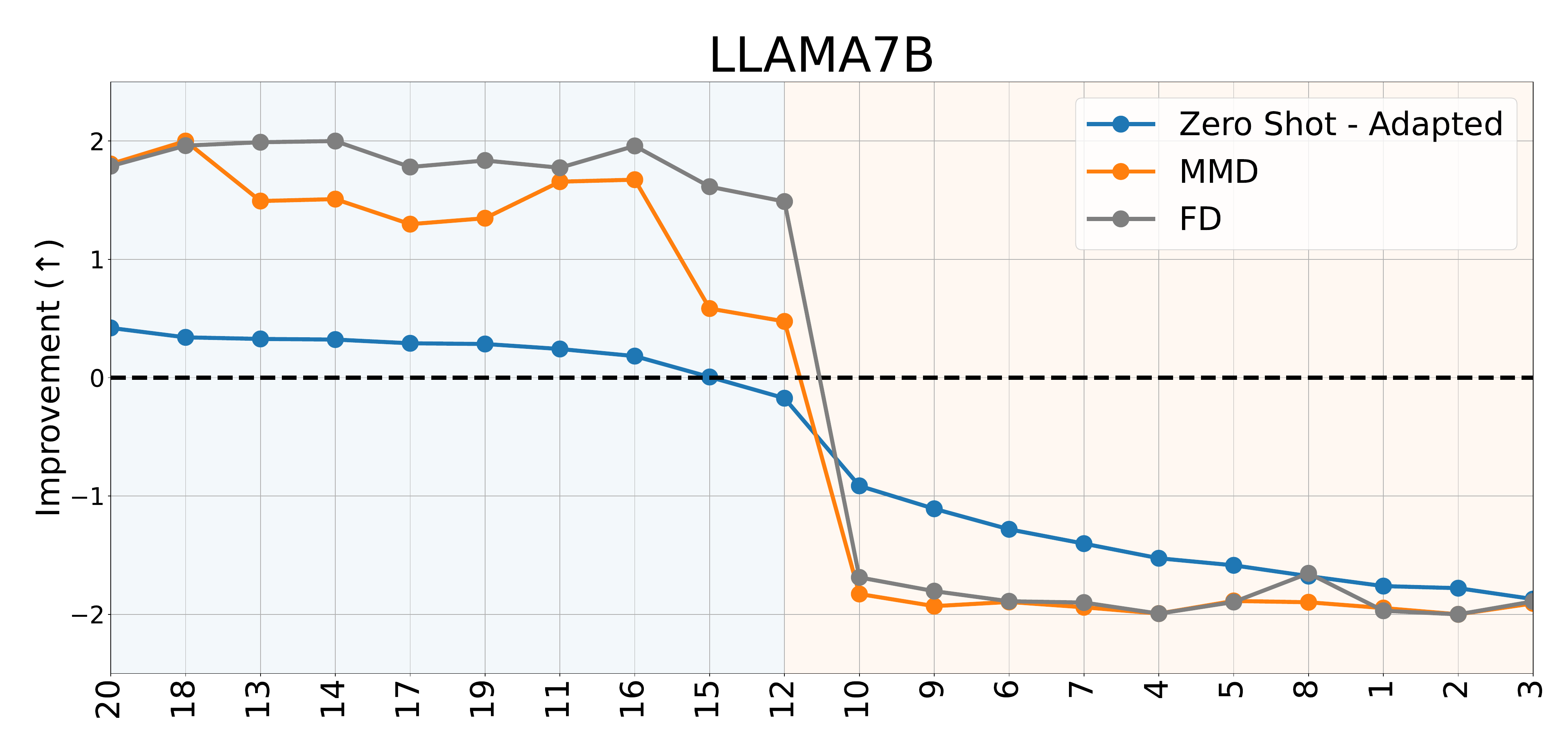}
	\caption{Perplexity change after adaptation {(denoted with Zero Shot - Adapted), where - stands for subtraction of perplexities} (blue) and similarity measures (orange and green, re-scaled for visualization purposes), plotted against adapted domains ($x$-axis), which are S2ORC (Blue Shaded Area) and Wiki (Orange Shaded Area). Adaptation domain names corresponding to the IDs on the $x$-axis are presented in Appendix~\ref{sec:adaptation-domain}. Above the black dashed line are the domains for which adaptation improved the test perplexity. Interestingly, we observe a degradation in Wiki domains. When the model capacity increases the gap between zero shot and adaptation becomes smaller.} 
    \label{fig:zs-ft}
\end{figure*}

\section{Results}
\label{sec:results}

For different sizes of \gpt as well as \olmo and \llama, we first compute the \textit{zero-shot perplexity} (Figure~\ref{fig:zs-ft}) on all domains. After the additional pretraining the models on each domain individually, we also compute the test perplexity on the corresponding test sets and refer it as the \textit{adaptation perplexity} (Figure~\ref{fig:zs-ft}). Because the models are tested on the same domain as they are trained on, one would naturally expect the perplexity to improve. However, our main findings in Figure~\ref{fig:zs-ft} present the opposite.

To demonstrate the degrading performance, in Figure~\ref{fig:zs-ft} we plot the difference between zero-shot and adaptation perplexities. For illustration purposes, we choose the most extreme 20 domains for each experiment, i.e., the ones on which the performance improves/degrades the most. The findings consistently show that adaptation improves perplexity for a subset of domains from the S2ORC portion of M2D2 while adaptation on the Wiki portion worsens the perplexity.

\paragraph{Domain similarity.} To understand potential causes for this, we check the similarity between the original corpora and adaptation domains. As shown in Figure~\ref{fig:zs-ft}, adaptation to Wiki domains, which are similar to original corpora, causes an increase in perplexity. We do not observe this strange phenomenon in S2ORC domains, where adaptation always improves perplexity.

\begin{figure*}[ht!]
    \centering
    \small GPT2-LARGE \\
    \includegraphics[width=0.99\linewidth]{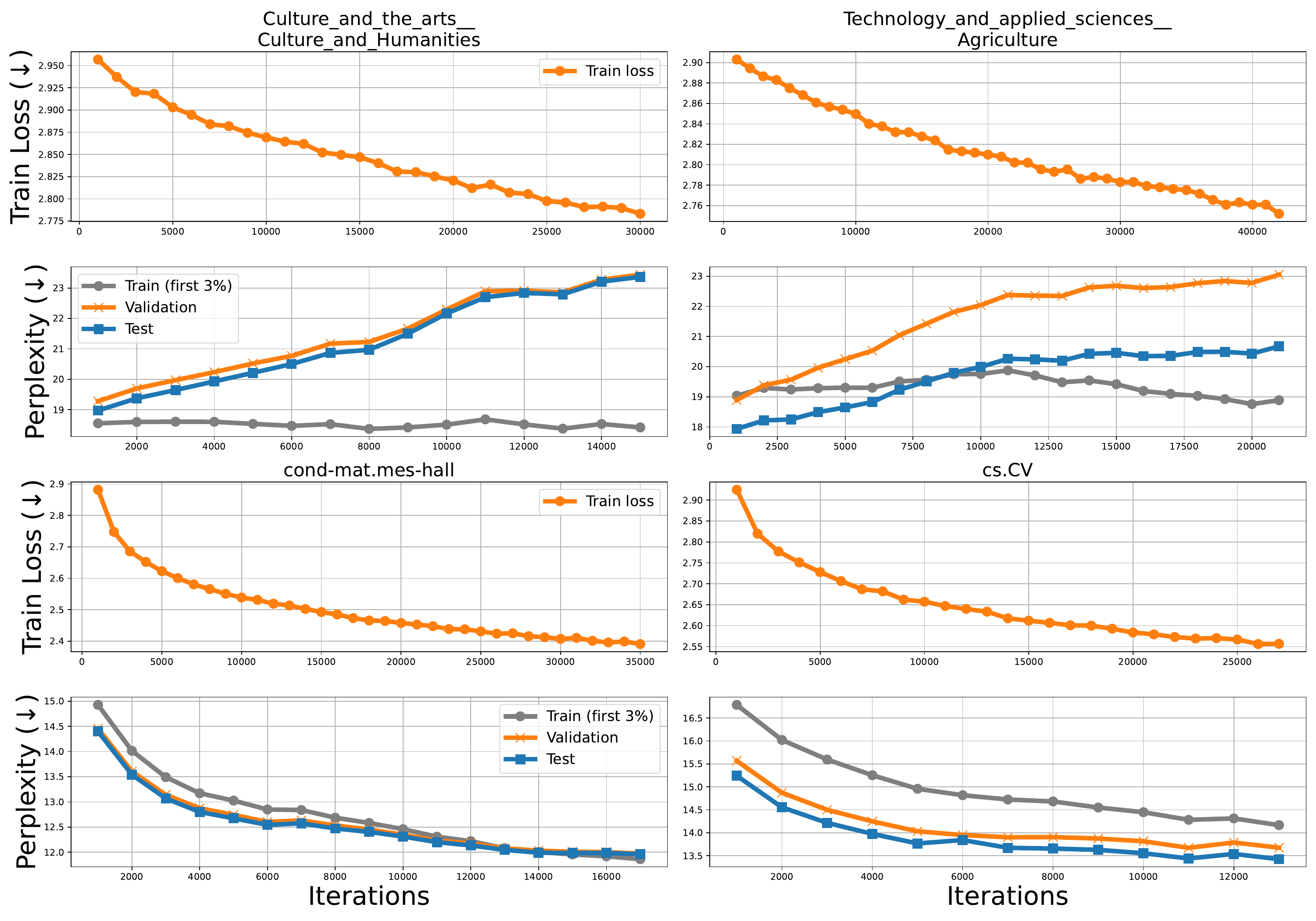}
    \caption{\label{fig:sgd} The perplexities computed on 4 domains during  pretraining. Note that we pretrain only for one epoch, i.e., the first 3\% of the training data is never seen again.}
\end{figure*}

\begin{figure*}[ht!]
    \centering
    \includegraphics[width=0.49\linewidth]{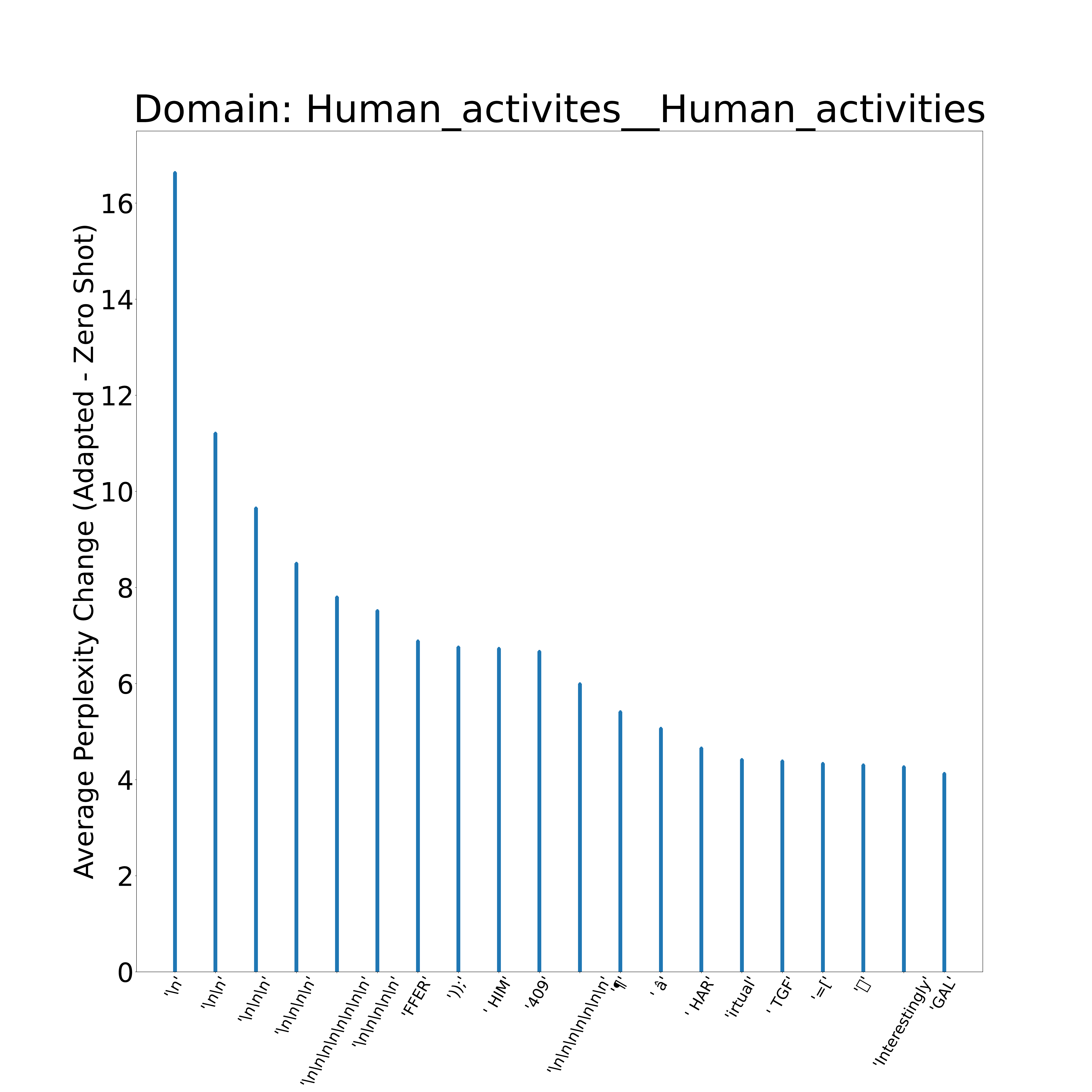}
    \includegraphics[width=0.49\linewidth]{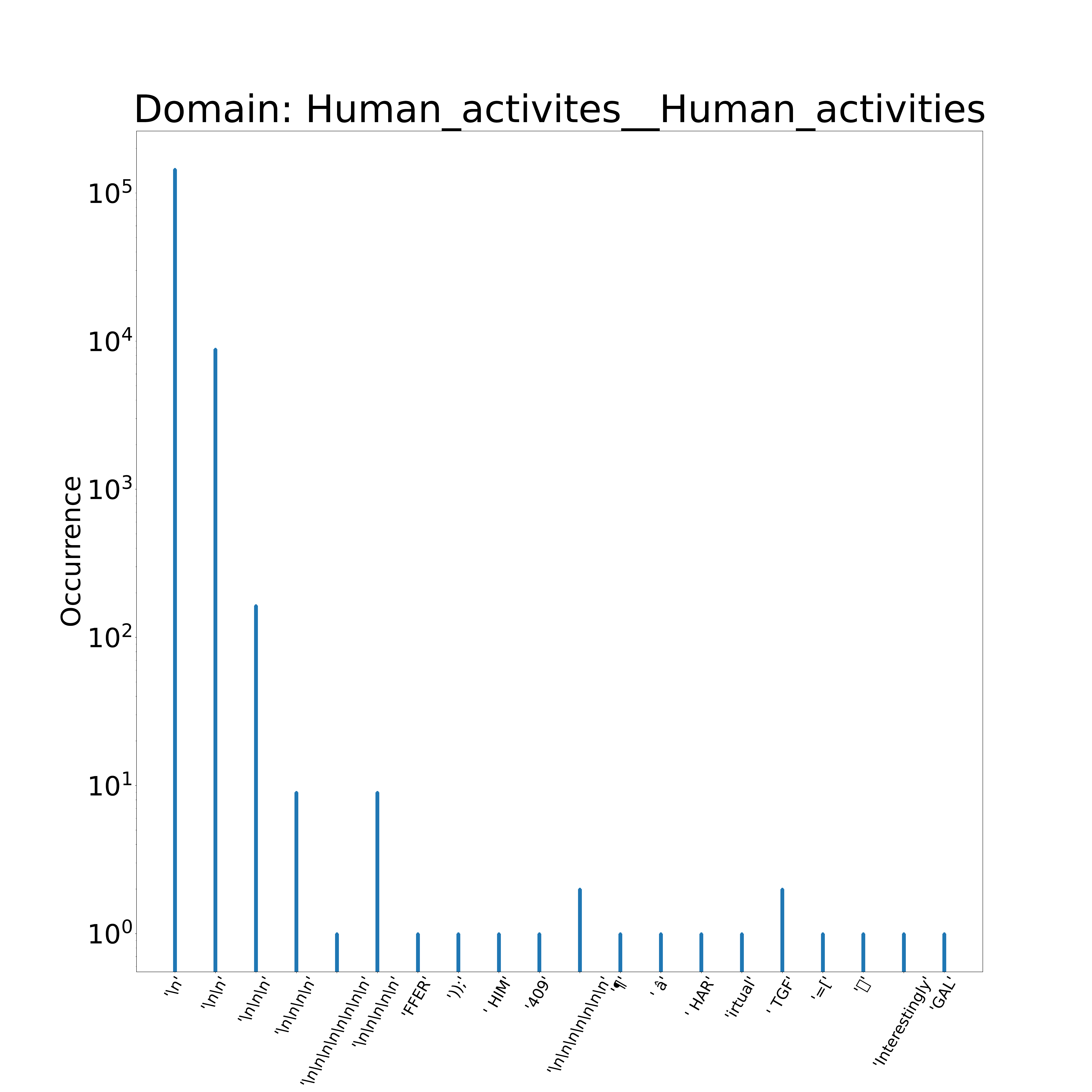}
    \caption{\label{fig:tokens} The figure on the left presents our token-level analysis of the \olmo model on the train portion of \textit{Human activities} subdomain of the Wiki corpus. The $x$-axis displays the tokens that exhibit the greatest increase in perplexity after domain adaptation, while the $y$-axis shows the corresponding average degradation in perplexity, which spans orders of magnitude. The figure on the right presents the occurrences of the tokens. Special tokens like ``\texttt{\textbackslash n}'' and ``\texttt{\textbackslash n\textbackslash n}'' are the most seen tokens. $y$-axis is in the log scale.}
\end{figure*}

\paragraph{What happens during gradient descent?} 
% \subsection{Generalization may deteriorate due to continual learning}
For \gpts and \gptl, we further visualize training curves on four randomly chosen domains in Figure~\ref{fig:sgd}. For all model sizes, the training and test losses on S2ORC domains steadily decrease, aligning with expectations. Interestingly, for certain domains such as Culture and Humanities or Agriculture, the loss computed on the validation set, test set as well as the first three percent of the training set increases during optimization.
In other words, while the model is optimized, its performance on the recently seen data as well as unseen data from the same data distribution deteriorates. 
Finally, we observe that an increased model capacity seems to help with this degradation.
% This counter-intuitive observation suggests that models pretrained on extensive, high-quality datasets possess better generalization capabilities than those further trained on specific Wiki portions.
\paragraph{Token-level observations.} 
Next, we dive deeper into our main finding by analyzing how adaptation changes the perplexity on all unique tokens. 
% the improving and degrading tokens in both improving and degrading domains' train and test splits. 
For this, we randomly sample 128 text chunks with 4096 tokens from the training and test set of adaptation domains, compute the perplexities of the zero-shot and adapted \olmo model, and group the perplexities by the ID of the predicted token. 
Most strikingly, we noticed that regardless of the training domain, adaptation significantly degrades the perplexity on special tokens such as ``\texttt{\textbackslash n}'' and ``\texttt{\textbackslash n\textbackslash n}'', which form a substantial subset of all tokens in the test set. Figure~\ref{fig:tokens} shows the top-20 tokens on which perplexity degrades the most when \olmo is adapted to \textit{Human activities} domain. 
To give further qualitative examples; we observed that training on \texttt{high energy physics} domain improves test perplexity on tokens such as ``\texttt{float}'', ``\texttt{asymptotic}'', and ``\texttt{projections}'' while perplexity on ``\texttt{specifically}'', ``\texttt{complete}'', and ``\texttt{string}'' get worse.

% This is an interesting finding as it loses general knowledge about what comes next after a paragraph. Also, for the improving domain, model increases it's performance on domain related tokens.

\section{Discussion}
\label{sec:discussion}
This short paper aims to improve our understanding of additional pretraining of models already pretrained on diverse text corpora. Interestingly, we observed that within-domain perplexity does not always increase. Below we summarize our main findings and takeaways:

\paragraph{Similarity between original corpus and adaptation domain affects the performance.}
When the original corpus and the adaptation domain are more similar, test perplexity in this adaptation domain after additional pretraining tends to increase. This phenomenon is not observed while adapting a less similar domain. Therefore, we recommend practitioners analyze the domains in their original pretraining corpora and then decide (not) to adapt.

\paragraph{Adaptation influences smaller models more}
Regardless of the training domain, the perplexity of \gpts models seems to change the most through adaptation. This finding suggests that adapting larger models may not be necessary.

\paragraph{Going beyond perplexity?}
Our token-level observations reveal that most of the perplexity degradation arises from specific special tokens. This indicates that perplexity alone may not fully capture the impact of adaptation. For future work, we plan to extend our analysis to include domain-specific tokens to better quantify the gains and degradations resulting from adaptation, providing recipes for when to stop adaptation or continue adapting.

% Still, existing work showed that domain adaptation still somewhat works.
% Domain adaptation is usually based on continued pretraining on a specific corpus. 
% Its promise is to obtain better experts, which intuitively makes sense.
% Yet, because we often don't know what constitutes training data, a model that was trained on a diverse corpus by engineers may already be an expert on the topic of interest.
% This is what we test in this work.

\section{Limitations}
\label{sec:limitations}
Our analyses require access to the training corpus of a pretrained LLM, thus not applicable to all models. One way to overcome this issue could be gathering a large representative corpus across the Internet and conducting analyses using this corpus. Further, our analysis quantifies the gains and degradations only via perplexity while computing downstream performance would be equally interesting.

\section{Acknowledgments}
This research was enabled in part by support provided by Calcul Québec and the Digital Research Alliance of Canada. Çağatay Yıldız and Matthias Bethge are members of the Machine Learning Cluster of Excellence, funded by the Deutsche Forschungsgemeinschaft (DFG, German Research Foundation) under Germany’s Excellence Strategy – EXC number 2064/1 – Project number 390727645. Matthia Bethge acknowledges financial support via the Open Philanthropy Foundation funded by the Good Ventures Foundation.

\newpage

\bibliography{main}

\begin{thebibliography}{15}
\providecommand{\natexlab}[1]{#1}

\bibitem[{Cheng et~al.(2023)Cheng, Huang, and Wei}]{cheng2023adapting}
Daixuan Cheng, Shaohan Huang, and Furu Wei. 2023.
\newblock Adapting large language models via reading comprehension.
\newblock \emph{arXiv preprint arXiv:2309.09530}.

\bibitem[{Computer(2023)}]{together2023redpajama}
Together Computer. 2023.
\newblock \href {https://github.com/togethercomputer/RedPajama-Data} {Redpajama: an open dataset for training large language models}.

\bibitem[{Dowson and Landau(1982)}]{frechet1982}
D.C Dowson and B.V Landau. 1982.
\newblock \href {https://doi.org/10.1016/0047-259X(82)90077-X} {The fréchet distance between multivariate normal distributions}.
\newblock \emph{Journal of Multivariate Analysis}, 12(3):450--455.

\bibitem[{Gokaslan and Cohen(2019)}]{Gokaslan2019OpenWeb}
Aaron Gokaslan and Vanya Cohen. 2019.
\newblock Openwebtext corpus.
\newblock \url{http://Skylion007.github.io/OpenWebTextCorpus}.

\bibitem[{Gretton et~al.(2012)Gretton, Borgwardt, Rasch, Sch{{\"o}}lkopf, and Smola}]{gretton12mmd}
Arthur Gretton, Karsten~M. Borgwardt, Malte~J. Rasch, Bernhard Sch{{\"o}}lkopf, and Alexander Smola. 2012.
\newblock \href {http://jmlr.org/papers/v13/gretton12a.html} {A kernel two-sample test}.
\newblock \emph{Journal of Machine Learning Research}, 13(25):723--773.

\bibitem[{Groeneveld et~al.(2024)Groeneveld, Beltagy, Walsh, Bhagia, Kinney, Tafjord, Jha, Ivison, Magnusson, Wang, Arora, Atkinson, Authur, Chandu, Cohan, Dumas, Elazar, Gu, Hessel, Khot, Merrill, Morrison, Muennighoff, Naik, Nam, Peters, Pyatkin, Ravichander, Schwenk, Shah, Smith, Strubell, Subramani, Wortsman, Dasigi, Lambert, Richardson, Zettlemoyer, Dodge, Lo, Soldaini, Smith, and Hajishirzi}]{groeneveld2024olmo}
Dirk Groeneveld, Iz~Beltagy, Pete Walsh, Akshita Bhagia, Rodney Kinney, Oyvind Tafjord, Ananya~Harsh Jha, Hamish Ivison, Ian Magnusson, Yizhong Wang, Shane Arora, David Atkinson, Russell Authur, Khyathi~Raghavi Chandu, Arman Cohan, Jennifer Dumas, Yanai Elazar, Yuling Gu, Jack Hessel, Tushar Khot, William Merrill, Jacob Morrison, Niklas Muennighoff, Aakanksha Naik, Crystal Nam, Matthew~E. Peters, Valentina Pyatkin, Abhilasha Ravichander, Dustin Schwenk, Saurabh Shah, Will Smith, Emma Strubell, Nishant Subramani, Mitchell Wortsman, Pradeep Dasigi, Nathan Lambert, Kyle Richardson, Luke Zettlemoyer, Jesse Dodge, Kyle Lo, Luca Soldaini, Noah~A. Smith, and Hannaneh Hajishirzi. 2024.
\newblock Olmo: Accelerating the science of language models.

\bibitem[{Gururangan et~al.(2020)Gururangan, Marasovi{\'c}, Swayamdipta, Lo, Beltagy, Downey, and Smith}]{gururangan2020don}
Suchin Gururangan, Ana Marasovi{\'c}, Swabha Swayamdipta, Kyle Lo, Iz~Beltagy, Doug Downey, and Noah~A Smith. 2020.
\newblock Don't stop pretraining: Adapt language models to domains and tasks.
\newblock \emph{arXiv preprint arXiv:2004.10964}.

\bibitem[{Kaplan et~al.(2020)Kaplan, McCandlish, Henighan, Brown, Chess, Child, Gray, Radford, Wu, and Amodei}]{kaplan2020scaling}
Jared Kaplan, Sam McCandlish, Tom Henighan, Tom~B Brown, Benjamin Chess, Rewon Child, Scott Gray, Alec Radford, Jeffrey Wu, and Dario Amodei. 2020.
\newblock Scaling laws for neural language models.
\newblock \emph{arXiv preprint arXiv:2001.08361}.

\bibitem[{Neyshabur et~al.(2017)Neyshabur, Bhojanapalli, McAllester, and Srebro}]{neyshabur2017exploring}
Behnam Neyshabur, Srinadh Bhojanapalli, David McAllester, and Nati Srebro. 2017.
\newblock Exploring generalization in deep learning.
\newblock \emph{Advances in neural information processing systems}, 30.

\bibitem[{Radford et~al.(2019)Radford, Wu, Child, Luan, Amodei, Sutskever et~al.}]{radford2019language}
Alec Radford, Jeffrey Wu, Rewon Child, David Luan, Dario Amodei, Ilya Sutskever, et~al. 2019.
\newblock Language models are unsupervised multitask learners.
\newblock \emph{OpenAI blog}, 1(8):9.

\bibitem[{Rasley et~al.(2020)Rasley, Rajbhandari, Ruwase, and He}]{rasley2020deepspeed}
Jeff Rasley, Samyam Rajbhandari, Olatunji Ruwase, and Yuxiong He. 2020.
\newblock Deepspeed: System optimizations enable training deep learning models with over 100 billion parameters.
\newblock In \emph{Proceedings of the 26th ACM SIGKDD International Conference on Knowledge Discovery \& Data Mining}, pages 3505--3506.

\bibitem[{Reid et~al.(2022)Reid, Zhong, Gururangan, and Zettlemoyer}]{reid2022m2d2}
Machel Reid, Victor Zhong, Suchin Gururangan, and Luke Zettlemoyer. 2022.
\newblock M2d2: A massively multi-domain language modeling dataset.
\newblock \emph{arXiv preprint arXiv:2210.07370}.

\bibitem[{Reimers and Gurevych(2019)}]{reimers-2019-sentence-bert}
Nils Reimers and Iryna Gurevych. 2019.
\newblock \href {http://arxiv.org/abs/1908.10084} {Sentence-bert: Sentence embeddings using siamese bert-networks}.
\newblock In \emph{Proceedings of the 2019 Conference on Empirical Methods in Natural Language Processing}. Association for Computational Linguistics.

\bibitem[{Soldaini et~al.(2024)Soldaini, Kinney, Bhagia, Schwenk, Atkinson, Authur, Bogin, Chandu, Dumas, Elazar, Hofmann, Jha, Kumar, Lucy, Lyu, Lambert, Magnusson, Morrison, Muennighoff, Naik, Nam, Peters, Ravichander, Richardson, Shen, Strubell, Subramani, Tafjord, Walsh, Zettlemoyer, Smith, Hajishirzi, Beltagy, Groeneveld, Dodge, and Lo}]{dolma}
Luca Soldaini, Rodney Kinney, Akshita Bhagia, Dustin Schwenk, David Atkinson, Russell Authur, Ben Bogin, Khyathi Chandu, Jennifer Dumas, Yanai Elazar, Valentin Hofmann, Ananya~Harsh Jha, Sachin Kumar, Li~Lucy, Xinxi Lyu, Nathan Lambert, Ian Magnusson, Jacob Morrison, Niklas Muennighoff, Aakanksha Naik, Crystal Nam, Matthew~E. Peters, Abhilasha Ravichander, Kyle Richardson, Zejiang Shen, Emma Strubell, Nishant Subramani, Oyvind Tafjord, Pete Walsh, Luke Zettlemoyer, Noah~A. Smith, Hannaneh Hajishirzi, Iz~Beltagy, Dirk Groeneveld, Jesse Dodge, and Kyle Lo. 2024.
\newblock \href {https://arxiv.org/abs/2402.00159} {{Dolma: An Open Corpus of Three Trillion Tokens for Language Model Pretraining Research}}.
\newblock \emph{arXiv preprint}.

\bibitem[{Touvron et~al.(2023)Touvron, Lavril, Izacard, Martinet, Lachaux, Lacroix, Rozière, Goyal, Hambro, Azhar, Rodriguez, Joulin, Grave, and Lample}]{touvron2023llama}
Hugo Touvron, Thibaut Lavril, Gautier Izacard, Xavier Martinet, Marie-Anne Lachaux, Timothée Lacroix, Baptiste Rozière, Naman Goyal, Eric Hambro, Faisal Azhar, Aurelien Rodriguez, Armand Joulin, Edouard Grave, and Guillaume Lample. 2023.
\newblock \href {https://arxiv.org/abs/2302.13971} {Llama: Open and efficient foundation language models}.
\newblock \emph{Preprint}, arXiv:2302.13971.

\end{thebibliography}
\appendix
\newpage
\section{Appendix}
\subsection{Adaptation Domain Names}
\label{sec:adaptation-domain}

\begin{table}[ht!]
\begin{tabular}{|c|c|}
\hline
\textbf{ID} & \textbf{Domain}                                                                                   \\ \hline
1           & \begin{tabular}[c]{@{}c@{}}Society\_and\_social\_sciences\\ Society\end{tabular}                  \\ \hline
2           & Technology\_and\_applied\_sciences                                                                \\ \hline
3           & \begin{tabular}[c]{@{}c@{}}Human\_activites\\ Human\_activities\end{tabular}                      \\ \hline
4           & \begin{tabular}[c]{@{}c@{}}Technology\_and\_applied\_sciences\\ Agriculture\end{tabular}          \\ \hline
5           & \begin{tabular}[c]{@{}c@{}}Culture\_and\_the\_arts\\ Culture\_and\_Humanities\end{tabular}        \\ \hline
6           & \begin{tabular}[c]{@{}c@{}}History\_and\_events\\ By\_period\end{tabular}                         \\ \hline
7           & \begin{tabular}[c]{@{}c@{}}Philosophy\_and\_thinking\\ Philosophy\end{tabular}                    \\ \hline
8           & \begin{tabular}[c]{@{}c@{}}Natural\_and\_physical\_sciences\\ Biology\end{tabular}                \\ \hline
9           & \begin{tabular}[c]{@{}c@{}}Philosophy\_and\_thinking\\ Thinking\end{tabular}                      \\ \hline
10          & \begin{tabular}[c]{@{}c@{}}General\_referece\\ Further\_research\_tools\_and\_topics\end{tabular} \\ \hline
11          & hep-ex                                                                                            \\ \hline
12          & hep-lat                                                                                           \\ \hline
13          & nucl-ex                                                                                           \\ \hline
14          & cond-mat.str-el                                                                                   \\ \hline
15          & nucl-th                                                                                           \\ \hline
16          & math.SG                                                                                           \\ \hline
17          & supr-con                                                                                          \\ \hline
18          & cond-mat.supr-con                                                                                 \\ \hline
19          & math.AG                                                                                           \\ \hline
20          & astro-ph.HE                                                                                       \\ \hline
\end{tabular}
\end{table}

\newpage
\subsection{All Domain Names}
\label{sec:all-domain}
\begin{table}[ht!]
\centering
\begin{tabular}{|c|c|}
\hline
\textbf{ID} & \textbf{Domain Name}                                                                                    \\ \hline
1           & Art                                                                                                     \\ \hline
2           & Culture\_and\_the\_arts                                                                                 \\ \hline
3           & \begin{tabular}[c]{@{}c@{}}Culture\_and\_the\_arts\\ Culture\_and\_Humanities\end{tabular}              \\ \hline
4           & \begin{tabular}[c]{@{}c@{}}Culture\_and\_the\_arts\\ Games\_and\_Toys\end{tabular}                      \\ \hline
5           & \begin{tabular}[c]{@{}c@{}}Culture\_and\_the\_arts\\ Mass\_media\end{tabular}                           \\ \hline
6           & \begin{tabular}[c]{@{}c@{}}Culture\_and\_the\_arts\\ Performing\_arts\end{tabular}                      \\ \hline
7           & \begin{tabular}[c]{@{}c@{}}Culture\_and\_the\_arts\\ Sports\_and\_Recreation\end{tabular}               \\ \hline
8           & \begin{tabular}[c]{@{}c@{}}Culture\_and\_the\_arts\\ The\_arts\_and\_Entertainment\end{tabular}         \\ \hline
9           & \begin{tabular}[c]{@{}c@{}}Culture\_and\_the\_arts\\ Visual\_arts\end{tabular}                          \\ \hline
10          & General\_referece                                                                                       \\ \hline
11          & \begin{tabular}[c]{@{}c@{}}General\_referece\\ Further\_research\_tools\_and\_topics\end{tabular}       \\ \hline
12          & \begin{tabular}[c]{@{}c@{}}General\_referece\\ Reference\_works\end{tabular}                            \\ \hline
13          & Health\_and\_fitness                                                                                    \\ \hline
14          & \begin{tabular}[c]{@{}c@{}}Health\_and\_fitness\\ Exercise\end{tabular}                                 \\ \hline
15          & \begin{tabular}[c]{@{}c@{}}Health\_and\_fitness\\ Health\_science\end{tabular}                          \\ \hline
16          & \begin{tabular}[c]{@{}c@{}}Health\_and\_fitness\\ Human\_medicine\end{tabular}                          \\ \hline
17          & \begin{tabular}[c]{@{}c@{}}Health\_and\_fitness\\ Nutrition\end{tabular}                                \\ \hline
18          & \begin{tabular}[c]{@{}c@{}}Health\_and\_fitness\\ Public\_health\end{tabular}                           \\ \hline
19          & \begin{tabular}[c]{@{}c@{}}Health\_and\_fitness\\ Self\_care\end{tabular}                               \\ \hline
20          & History\_and\_events                                                                                    \\ \hline
21          & \begin{tabular}[c]{@{}c@{}}History\_and\_events\\ By\_continent\end{tabular}                            \\ \hline
22          & \begin{tabular}[c]{@{}c@{}}History\_and\_events\\ By\_period\end{tabular}                               \\ \hline
23          & \begin{tabular}[c]{@{}c@{}}History\_and\_events\\ By\_region\end{tabular}                               \\ \hline
24          & Human\_activites                                                                                        \\ \hline
25          & \begin{tabular}[c]{@{}c@{}}Human\_activites\\ Human\_activities\end{tabular}                            \\ \hline
26          & \begin{tabular}[c]{@{}c@{}}Human\_activites\\ Impact\_of\_human\_activity\end{tabular}                  \\ \hline
27          & Mathematics\_and\_logic                                                                                 \\ \hline

\end{tabular}
\end{table}

\begin{table}[]
\begin{tabular}{|c|c|}
\hline
\textbf{ID} & \textbf{Domain Name}   
\\ \hline
28          & \begin{tabular}[c]{@{}c@{}}Mathematics\_and\_logic\\ Fields\_of\_mathematics\end{tabular}               \\ \hline
29          & \begin{tabular}[c]{@{}c@{}}Mathematics\_and\_logic\\ Logic\end{tabular}                                 \\ \hline
30          & \begin{tabular}[c]{@{}c@{}}Mathematics\_and\_logic\\ Mathematics\end{tabular}                           \\ \hline
31          & Natural\_and\_physical\_sciences                                                                        \\ \hline
32          & \begin{tabular}[c]{@{}c@{}}Natural\_and\_physical\_sciences\\ Biology\end{tabular}                      \\ \hline
33          & \begin{tabular}[c]{@{}c@{}}Natural\_and\_physical\_sciences\\ Earth\_sciences\end{tabular}              \\ \hline
34          & \begin{tabular}[c]{@{}c@{}}Natural\_and\_physical\_sciences\\ Nature\end{tabular}                       \\ \hline
35          & \begin{tabular}[c]{@{}c@{}}Natural\_and\_physical\_sciences\\ Physical\_sciences\end{tabular}           \\ \hline
36          & Philosophy                                                                                              \\ \hline
37          & Philosophy\_and\_thinking                                                                               \\ \hline
38          & \begin{tabular}[c]{@{}c@{}}Philosophy\_and\_thinking\\ Philosophy\end{tabular}                          \\ \hline
39          & \begin{tabular}[c]{@{}c@{}}Philosophy\_and\_thinking\\ Thinking\end{tabular}                            \\ \hline
40          & Religion\_and\_belief\_systems                                                                          \\ \hline
41          & \begin{tabular}[c]{@{}c@{}}Religion\_and\_belief\_systems\\ Allah\end{tabular}                          \\ \hline
42          & \begin{tabular}[c]{@{}c@{}}Religion\_and\_belief\_systems\\ Belief\_systems\end{tabular}                \\ \hline
43          & \begin{tabular}[c]{@{}c@{}}Religion\_and\_belief\_systems\\ Major\_beliefs\_of\_the\_world\end{tabular} \\ \hline
44          & Society\_and\_social\_sciences                                                                          \\ \hline
45          & \begin{tabular}[c]{@{}c@{}}Society\_and\_social\_sciences\\ Social\_sciences\end{tabular}               \\ \hline
46          & \begin{tabular}[c]{@{}c@{}}Society\_and\_social\_sciences\\ Society\end{tabular}                        \\ \hline
47          & Technology\_and\_applied\_sciences                                                                      \\ \hline
48          & \begin{tabular}[c]{@{}c@{}}Technology\_and\_applied\_sciences\\ Agriculture\end{tabular}                \\ \hline
49          & \begin{tabular}[c]{@{}c@{}}Technology\_and\_applied\_sciences\\ Computing\end{tabular}                  \\ \hline
50          & \begin{tabular}[c]{@{}c@{}}Technology\_and\_applied\_sciences\\ Engineering\end{tabular}                \\ \hline
51          & \begin{tabular}[c]{@{}c@{}}Technology\_and\_applied\_sciences\\ Transport\end{tabular}                  \\ \hline
52          & astro-ph.CO                                                                                             \\ \hline
53          & astro-ph.EP                                                                                             \\ \hline
54          & astro-ph.HE                                                                                             \\ \hline
55          & astro-ph.IM                                                                                             \\ \hline
56          & astro-ph.SR                                                                                             \\ \hline
57          & atom-ph                                                                                                 \\ \hline
58          & chem-ph                                                                                                 \\ \hline
\end{tabular}
\end{table}

\begin{table}[ht!]
\centering
\begin{tabular}{|c|c|}
\hline
\textbf{ID} & \textbf{Domain Name}   

\\ \hline

59          & cond-mat.dis-nn                                                                                         \\ \hline
60          & cond-mat.mes-hall                                                                                       \\ \hline
61          & cond-mat.mtrl-sci                                                                                       \\ \hline
62          & cond-mat.other                                                                                          \\ \hline
63          & cond-mat.quant-gas                                                                                      \\ \hline
64          & cond-mat.soft                                                                                           \\ \hline
65          & cond-mat.stat-mech                                                                                      \\ \hline
66          & cond-mat.str-el                                                                                         \\ \hline
67          & cond-mat.supr-con                                                                                       \\ \hline
68          & cs.AI                                                                                                   \\ \hline
69          & cs.AR                                                                                                   \\ \hline
70          & cs.CC                                                                                                   \\ \hline
71          & cs.CE                                                                                                   \\ \hline
72          & cs.CG                                                                                                   \\ \hline
73          & cs.CL                                                                                                   \\ \hline
74          & cs.CR                                                                                                   \\ \hline
75          & cs.CV                                                                                                   \\ \hline
76          & cs.CY                                                                                                   \\ \hline
77          & cs.DB                                                                                                   \\ \hline
78          & cs.DC                                                                                                   \\ \hline
79          & cs.DL                                                                                                   \\ \hline
80          & cs.DM                                                                                                   \\ \hline
81          & cs.DS                                                                                                   \\ \hline
82          & cs.ET                                                                                                   \\ \hline
83          & cs.FL                                                                                                   \\ \hline
84          & cs.GL                                                                                                   \\ \hline
85          & cs.GR                                                                                                   \\ \hline
86          & cs.GT                                                                                                   \\ \hline
87          & cs.HC                                                                                                   \\ \hline
88          & cs.IR                                                                                                   \\ \hline
89          & cs.LG                                                                                                   \\ \hline
90          & cs.LO                                                                                                   \\ \hline
91          & cs.MA                                                                                                   \\ \hline
92          & cs.MM                                                                                                   \\ \hline
93          & cs.MS                                                                                                   \\ \hline
94          & cs.NA                                                                                                   \\ \hline
95          & cs.NE                                                                                                   \\ \hline
96          & cs.NI                                                                                                   \\ \hline
97          & cs.OH                                                                                                   \\ \hline
98          & cs.OS                                                                                                   \\ \hline
99          & cs.PF        
 \\ \hline
 100         & cs.PL                                                                                                   \\ \hline
101         & cs.RO                                                                                                   \\ \hline
102         & cs.SC                                                                                                   \\ \hline
103         & cs.SD                                                                                                   \\ \hline
104         & cs.SE                                                                                                   \\ \hline
105         & cs.SI                                                                                                   \\ \hline
106         & cs.SY                                                                                                   \\ \hline

\end{tabular}
\end{table}

\begin{table}[h]
\centering
\begin{tabular}{|c|c|}
\hline
\textbf{ID} & \textbf{Domain Name}   
                                                                               \\ \hline

107         & econ.EM                                                                                                 \\ \hline
108         & econ.TH                                                                                                 \\ \hline
109         & eess.AS                                                                                                 \\ \hline
110         & eess.IV                                                                                                 \\ \hline
111         & eess.SP                                                                                                 \\ \hline
112         & gr-qc                                                                                                   \\ \hline
113         & hep-ex                                                                                                  \\ \hline
114         & hep-lat                                                                                                 \\ \hline
115         & math.AC                                                                                                 \\ \hline
116         & math.AG                                                                                                 \\ \hline
117         & math.AP                                                                                                 \\ \hline
118         & math.AT                                                                                                 \\ \hline
119         & math.CA                                                                                                 \\ \hline
120         & math.CT                                                                                                 \\ \hline
121         & math.CV                                                                                                 \\ \hline
122         & math.DG                                                                                                 \\ \hline

123         & math.DS                                                                                                 \\ \hline
124         & math.FA                                                                                                 \\ \hline
125         & math.GM                                                                                                 \\ \hline
126         & math.GN                                                                                                 \\ \hline
127         & math.GR                                                                                                 \\ \hline
128         & math.GT                                                                                                 \\ \hline
129         & math.HO                                                                                                 \\ \hline
130         & math.KT                                                                                                 \\ \hline
131         & math.LO                                                                                                 \\ \hline
132         & math.MG                                                                                                 \\ \hline
133         & math.NA                                                                                                 \\ \hline
134         & math.NT                                                                                                 \\ \hline
135         & math.OA                                                                                                 \\ \hline
136         & math.OC                                                                                                 \\ \hline
137         & math.PR                                                                                                 \\ \hline
138         & math.QA                                                                                                 \\ \hline
139         & math.RA                                                                                                 \\ \hline
140         & math.RT                                                                                                 \\ \hline
141         & math.SG                                                                                                 \\ \hline
142         & math.SP                                                                                                 \\ \hline
143         & nlin.AO                                                                                                 \\ \hline
144         & nlin.CD                                                                                                 \\ \hline
145         & nlin.CG                                                                                                 \\ \hline
146         & nlin.PS                                                                                                 \\ \hline
147         & nlin.SI                                                                                                 \\ \hline
148         & nucl-ex                                                                                                 \\ \hline
149         & nucl-th                                                                                                 \\ \hline
150         & physics.acc-ph                                                                                          \\ \hline
151         & physics.ao-ph                                                                                           \\ \hline
152         & physics.app-ph                                                                                          \\ \hline
153         & physics.atm-clus                                                                                        \\ \hline
154         & physics.atom-ph                                                                                         \\ \hline
155         & physics.bio-ph                                                                                          \\ \hline
156         & physics.chem-ph                                                                                         \\ \hline
157         & physics.class-ph                                                                                        \\ \hline
\end{tabular}
\end{table}

\begin{table}[h]
\centering
\begin{tabular}{|c|c|}
\hline
\textbf{ID} & \textbf{Domain Name}   
\\ \hline

158         & physics.comp-ph                                                                                         \\ \hline
159         & physics.data-an                                                                                         \\ \hline
160         & physics.ed-ph                                                                                           \\ \hline
161         & physics.flu-dyn                                                                                         \\ \hline
162         & physics.gen-ph                                                                                          \\ \hline
163         & physics.geo-ph                                                                                          \\ \hline
164         & physics.hist-ph                                                                                         \\ \hline
165         & physics.ins-det                             \\ 
\hline       
166         & physics.med-ph                                                                                          \\ \hline
167         & physics.optics                                                                                          \\ \hline
168         & physics.plasm-ph                                                                                        \\ \hline
169         & physics.pop-ph                                                                                          \\ \hline

170         & physics.soc-ph                                                                                          \\ \hline
171         & physics.space-ph                                                                                        \\ \hline
172         & plasm-ph                                                                                                \\ \hline
173         & q-bio                                                                                                   \\ \hline
174         & q-bio.BM                                                                                                \\ \hline
175         & q-bio.CB                                                                                                \\ \hline
176         & q-bio.GN                                                                                                \\ \hline
177         & q-bio.MN                                                                                                \\ \hline
178         & q-bio.NC                                                                                                \\ \hline
179         & q-bio.OT                                                                                                \\ \hline
180         & q-bio.PE                                                                                                \\ \hline
181         & q-bio.QM                                                                                                \\ \hline
182         & q-bio.SC                                                                                                \\ \hline
183         & q-bio.TO                                                                                                \\ \hline
184         & q-fin.CP                                                                                                \\ \hline
185         & q-fin.EC                                                                                                \\ \hline
186         & q-fin.GN                                                                                                \\ \hline
187         & q-fin.MF                                                                                                \\ \hline
188         & q-fin.PM                                                                                                \\ \hline
189         & q-fin.PR                                                                                                \\ \hline
190         & q-fin.RM                                                                                                \\ \hline
191         & q-fin.ST                                                                                                \\ \hline
192         & q-fin.TR                                                                                                \\ \hline
193         & quant-ph                                                                                                \\ \hline
194         & stat.AP                                                                                                 \\ \hline
195         & stat.CO                                                                                                 \\ \hline
196         & stat.ME                                                                                                 \\ \hline
197         & stat.ML                                                                                                 \\ \hline
198         & stat.OT                                                                                                 \\ \hline
199         & supr-con                                                                                                \\ \hline

\end{tabular}
\end{table}

\newpage
\begin{figure*}[b]
	\centering
	\includegraphics[width=\linewidth]{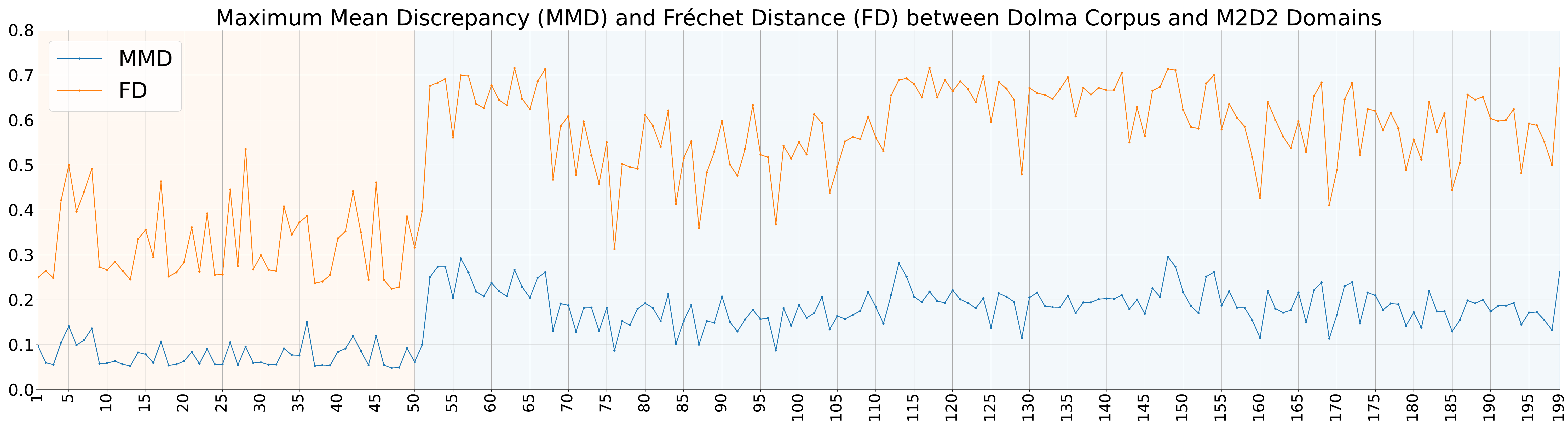}
	\caption{Domain IDs ($x$ axis). MMD and FD scores between Dolma and M2D2  Domains ($y$ axis). Wiki (blue shaded area) portion is closer to source corpora compared to the S2ORC (orange shaded area)  portion. Domain names are presented in Appendix~\ref{sec:all-domain}}.
	\label{fig:mmd-fid-dolma}
 
\end{figure*}

\begin{figure*}[b]
	\centering
	\includegraphics[width=\linewidth]{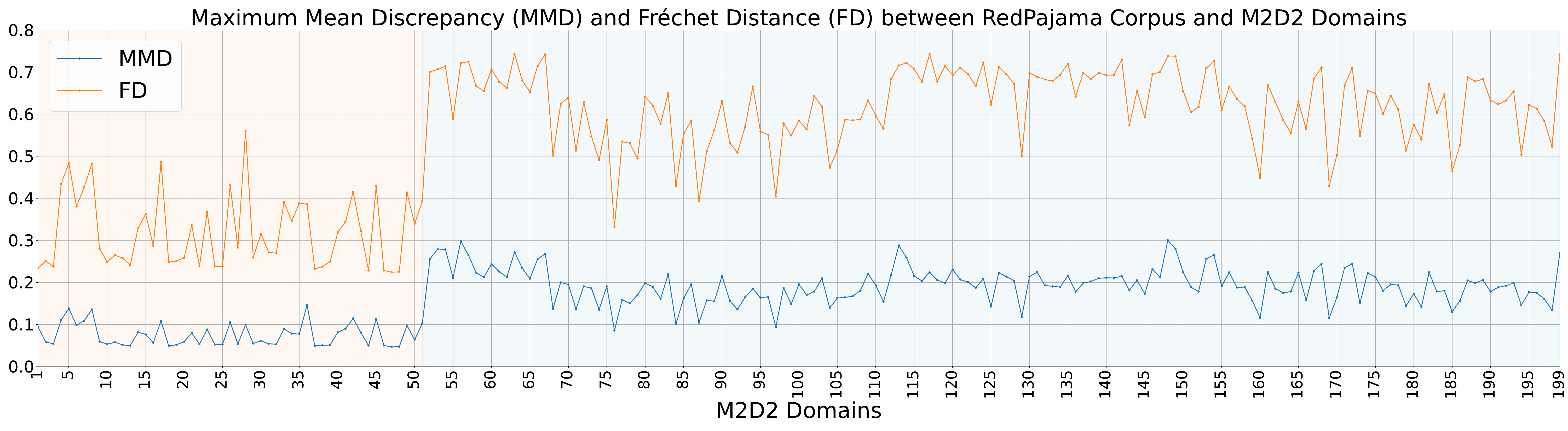}
	\caption{Domain IDs ($x$ axis). MMD and FD scores between RedPajama and M2D2  Domains ($y$ axis). Wiki (blue shaded area) portion is closer to source corpora compared to the S2ORC (orange shaded area)  portion. Domain names are presented in Appendix~\ref{sec:all-domain}}.
	\label{fig:mmd-fid-dolma}
 
\end{figure*}

\end{document}